\documentclass[11pt]{article}

\usepackage[preprint]{acl}
\usepackage[T1]{fontenc}
\usepackage[utf8]{inputenc} 
\usepackage[T1]{fontenc} 
\usepackage[para]{threeparttable}
\usepackage{hyperref}
\usepackage{url}
\usepackage{booktabs} 
\usepackage[colorinlistoftodos]{todonotes}
\usepackage{algorithmic}
\usepackage[linesnumbered, ruled, vlined,boxed]{algorithm2e}
\usepackage{multirow}
\usepackage{tablefootnote}
\usepackage{graphicx}
\usepackage{subfigure}
\usepackage{flushend}
\usepackage{color}
\usepackage{enumitem}
\usepackage{makecell}
\usepackage{footmisc}

\usepackage{xcolor}
\usepackage{xspace}
\usepackage{arydshln}

\newcommand{\xz}[1]{\textbf{\color{blue}[(Xu: #1 )]}}
\newcommand{\method}{\textbf{BIPro}\xspace}

\newcommand{\hide}[1]{}

\usepackage[]{threeparttable}

\newcommand{\vpara}[1]{\vspace{0.07in}\noindent\textbf{#1 }}
\newcommand{\beq}[1]{\vspace{-0.07in}\begin{equation}#1\end{equation}}

\begin{document}


\title{BIPro: Zero-shot Chinese Poem Generation via Block Inverse Prompting Constrained Generation Framework}
\author{Xu Zou \\
Zhipu. AI \\
\texttt{xz\_mailbox@xuzou.cn} }

\maketitle
\begin{abstract}
Recently, generative pre-trained models have made significant strides, particularly highlighted by the release of ChatGPT and GPT-4, which exhibit superior cross-domain capabilities. 
However, these models still face challenges on constrained writing tasks like poem generation under open-domain titles. 

In response to this challenge, we introduce \textbf{B}lock \textbf{I}nverse \textbf{Pro}mpting (\textbf{BIPro}) constrained generation framework. \textbf{BIPro} leverages two block inverse prompting methods, revise and rewrite, that mimic the process of human text writing using block generative models. It significantly improves the zero-shot generation quality on the formidable constrained generation task of open-domain traditional-form Chinese poem generation. 

Based on a less powerful block generative model GLM-10B-Chinese, poems composed via \textbf{BIPro} without priming or additional training outperform both most advanced direct generative systems like GPT-4 or GLM-4 and best domain-specific systems such as Yusheng, Shisanbai, or Baidu Poetry Helper in human evaluation by proficient poets.

Finally, \textbf{BIPro} considerably narrows the gap between AI-generated works and short-listed human literary arts in another human evaluation, unveiling the promising potential of block generative models in improving the quality of constrained generation. 
\end{abstract}

%
\section{Introduction}\label{sec:intro}
\begin{figure*}
\includegraphics[width=1.05\textwidth]{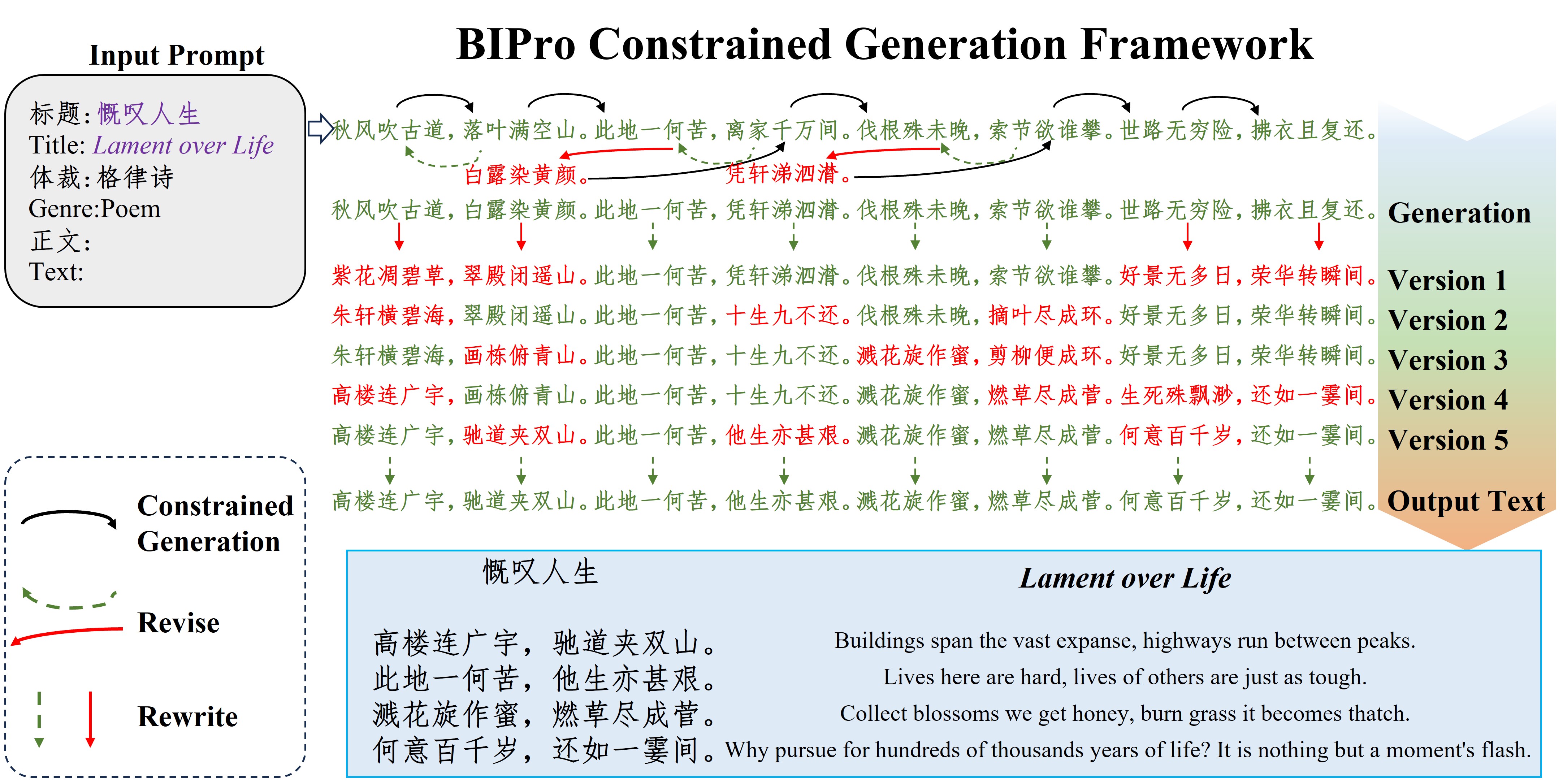}
\caption{\label{fig:poemgen}  The generation process of poem ``\textit{Lament over Life}" under \method framework. Sentences are generated with constraints using block generative model. Each sentence is revised after its subsequent sentence is generated. The full poem endures several rounds of rewrite. }
\vspace{-0.1in}
\end{figure*}
The current decade is marked by remarkable advancements in the field of generative pre-trained models. OpenAI's recent releases, ChatGPT and GPT-4~\cite{achiam2023gpt}, have become standout performers across a range of generative tasks, including translation, article writing, problem-solving, code generation, and image creation.

These advanced models are swiftly adopted in numerous social sectors, and AI-generated content is permeating our daily lives.~\cite{du2023effect,baldassarre2023social}. 

Constrained writing is a literary technique in which the author is bounded by some condition that imposes a certain pattern, often enhancing the aesthetic merit of the text. The most well-known application of constrained writing is poetry, where constraints like rhyme or meter are usually applied. Poets who master those constraints, such as Li Bai or William Shakespeare, are sometimes regarded as icons of their civilizations.~\cite{xie2019shakespeare}

However, the very constraints that elevate the artistic value of texts also introduce significant challenges, as they limit the freedom of expression, demanding more deliberate and planned creation. Masterpieces of constrained writing typically emerge from multiple revisions. Authors think deeply before penning their words and produce numerous drafts, trying to find the ideal expression.

This process could elucidate why generative models like GPTs struggle in this domain~\cite{garbacea2022constrained}. Direct generative models sequentially produce tokens through autoregression, considering only preceding text and lacking the ability to revise what has already been generated. Although GPT-generated poems are almost indistinguishable from human masterpieces for general public~\cite{deng2024can}, they are not as good in the view of reviewers with expertise in the domain~\cite{sawicki2023bits}.

Inverse prompting~\cite{zou2021controllable} is a text generation method designed to improve generation quality by searching the best generation using perplexity of the inverse form of natural language as scorer. 
One of the key limitations of inverse prompting lies in its dependence on the existence of precise inverse forms to convey the same meaning, unable to handle cases where inverse forms are absent or imprecise.

In this paper, we explore how inverse prompting can be improved through integration with block generative models, models that enable intermediate text generation according to both preceding and subsequent context.  We introduce two novel block inverse prompting methods and establish a \textbf{B}lock \textbf{I}nverse \textbf{Pro}mpting (\method) framework for constrained generation. 

We implement our proposed \method framework on one of the most challenging constrained generation tasks, the open-domain traditional-form Chinese poem generation. 
Figure ~\ref{fig:poemgen} illustrates an example of the process to generate a poem under open-domain title ``\textit{Lament over Life}" using \method. 
Besides direct constrained generation, each sentence of the poem is revised after its subsequent sentence is generated. After the initial generation, the poem is then rewritten for multiple times, mimicking the way humans produce poems. Each rewrite yields better expressions and improves the quality of the poem.

The exemplary open-domain traditional-form Chinese poem in figure ~\ref{fig:poemgen} does not emerge from most advanced direct generative systems or specialized systems extensively trained on domain-specific data. It is created using a relatively weak model, GLM-10B~\cite{du2022glm}, as base model.
Although this model is outperformed by cutting-edge direct generative systems like GPT-4~\cite{achiam2023gpt} or GLM-4 in direct generation, and lacks domain-specific expertise compared with domain-specific systems like Yusheng~\cite{ma2023yu} or Shisanbai, \method leverages its unique advantage of intermediate text generation, and empowers it to craft poetry of unparalleled excellence.

Reviews of human poets demonstrate that the \method framework significantly improves the ability of traditional-form Chinese poem generation of GLM-10B. 
 Poems produced by \method framework have outperformed a variety of baselines, including best domain-specific approaches Yusheng or Shisanbai as well as leading direct generation systems like GPT-4 or GLM-4. \method narrows the gap between AI generated poems and short-listed human poems in \textit{Daily Poem} section on \textit{China Poetry} website.

To summarize, the paper mainly presents the following key contributions:
\begin{itemize}
\setlength{\itemsep}{-4pt}
\item We introduce \method framework to harness the distinctive capabilities of block generative models, allowing them to refine and improve generated content autonomously on constrained generation tasks.
\item \method framework significantly improves the quality of the generated texts,  enabling the less advanced block generative model GLM-10B to outperform both superior generative systems and domain-specific systems in creating open-domain traditional-form Chinese poetry.
\item The efficacy of the \method framework highlights the untapped potential of block generative models in producing high-quality constrained generations. 
\end{itemize}

\section{Related Works} \label{sec:related}
\vspace{-0.02in}
\subsection{Generative Pre-trained Language Models}
 Pre-training is a long-lasting concept in psychology and education.~\cite{parker1963magnitude}
It is first introduced to handle natural language via word embeddings~\cite{mikolov2013distributed}. Following works like BERT~\cite{devlin2018bert} and GPT~\cite{radford2018improving} expand it to transformer-based language models.

The commercial success of ChatGPT triggers a great wave of generative pre-trained models. 
Following ChatGPT, various generative pre-trained models are released within a short period of time. Well-known examples include GPT-4~\cite{achiam2023gpt},
LLaMA~\cite{touvron2023llama,touvron2023llama2}, Qwen~\cite{bai2023qwen}, GLM-4~\footnote{\url{https://chatglm.cn/}}, Falcon~\cite{almazrouei2023falcon}, Baichuan~\cite{yang2023baichuan}, ERNIE Bot~\cite{ernie} and Gimini~\cite{team2023gemini}. 
\vspace{-0.1in}
\subsection{Block Generative Models}
Generative pre-trained models typically produce text via direct auto-regressive generation, where each token is generated based on solely the preceding tokens.  Once generated, the new token is incorporated into the input sequence to facilitate the generation of the subsequent tokens.

GLM~\cite{du2022glm} is a departure from this trend as a block generative model that enables non-monotonic generation. It can generate middle texts of any length according to both previous and following texts using its unique block attention mechanism. However, its direct generation performance is not as impressive as direct generative models. As a result, subsequent iterations of the GLM series, ChatGLM and GLM-4 abandon the block generative designation. 
\vspace{-0.1in}
\subsection{Constrained Writing and Traditional-form Chinese Poem Generation}
\hide{
\begin{table}
   \centering
   \caption{\label{tab:methods} Table of related models and systems}
   \vspace{-0.1in}
   \begin{tabular}{ccc}
    \toprule
    Type & System & Domain \\
     \midrule
      \makecell[c]{Block \\Generation Model}&\makecell[c]{GLM-10B-\\Chinese} & general Chinese \\
     \midrule
     \multirow{5}*
    {\makecell[c]{Direct \\Generation System}}& GPT-4  & \multirow{5}*{general language}\\
     &GLM-4 & \\
     &Qwen &\\
     & ERNIE Bot & \\
     &Gimini & \\
     \midrule
    \multirow{4}*{\makecell[c]{Domain-specific \\Generation System}}  & \makecell[c]{Baidu \\Poetry Helper} & \makecell[c]{general \\ Chinese poetry}\\
    \cmidrule{2-3}
      & Jiuge  & \multirow{3}*{\makecell[c]{traditional-form \\Chinese poetry}}\\
     &Yusheng &\\
     & Shisanbai &  \\
     \midrule
     \makecell[c]{Constrained \\Generation Framework}& \makecell[c]{\method} & \makecell[c]{constrained\\ generation tasks }\\
     \bottomrule
  \end{tabular}
  \vspace{-0.1in}
\end{table}
}
Constrained writing is a writing scenario that the process of writing is bounded by constraints like limited vocabulary, rhyme, meter, usage of vowels, or other constraints. It is particularly challenging for neural language models~\cite{garbacea2022constrained}. In some constrained generation tasks, like word puzzles, the difficulty lies on finding a solution to satisfy hard constraints. In other tasks, the constraints themselves are not hard, but the target is to write as good text as possible under the constraints. These tasks are more challenging for neural language models, as they have to balance between constraints and qualities of generated texts. 

One of the most well-known applications of constrained writing is poetry. The task of traditional-form Chinese poem generation is one of the most prestigious. Many Chinese language models including Baichuan~\cite{yang2023baichuan}, GLM-4,and Qwen~\cite{bai2023qwen} highlight poem generation as one of their spotlights in their model applications. There is also a poem-specific Baidu Poetry Helper derived from ERNIE Bot~\cite{ernie} specialized on general poem generation.

Originally codified in the 13th century, the Pingshui~\cite{Nie1982Pingshui} rhyme scheme serves as a comprehensive set of rules governing the structure of traditional-form Chinese poems. It is widely-accepted as the standard for traditional-form Chinese poems. 

Being a time-honored task, there exists lots of domain-specific systems specified on creating traditional-form Chinese poems. The most famous instance is Jiuge~\cite{zhipeng2019jiuge}, while Shisanbai~\footnote{\url{https://www.aichpoem.net/\#/shisanbai/poem}} and Yusheng~\cite{ma2023yu} are better and more recent instances.

\hide{
Table ~\ref{tab:methods} offers a brief list of the types and domains of different generation systems mentioned in this paper. 

However, sometimes a bad expression of a single sentence may be especially harmful under constrained generation. It may not violate constraints instantly but adds huge difficulty for generation of following sentences. Under the case of monotonic generation, there's hardly any way to aggregate information of the following sentences to change already generated sentences. 

In this work, we focus on the Chinese version of GLM-10B~\cite{du2022glm}, a 9.87-billion open-source parameter non-monotonic pre-trained language model dated back to 2021.
Although GLM-10B is much behind on benchmark performances of direct generation or fine-tuned downstream tasks compared with second generation monotonic models, the invited experts' evaluation shows that the unique advantages of non-monotonic models can help it generate the best traditional Chinese poem under the proposed constrained generation framework without any additional domain-specific data. 
}

\section{Methodology}\label{sec:problem}
\hide{
\subsection{Notations}
\xz{Remove this section?}
\begin{table}[t]
   \centering
   \caption{\label{tab:notations} Notations }
   \begin{tabular}{p{0.65in}|p{1.75in}}
    \toprule
     Notation & Description \\
     \midrule
     $\mathcal{M}$ & generative language model \\
     $\mathcal{D}$ & probability distribution of tokens predicted by the model \\
     $t_p$ & prompt text\\
     $t_g$ & generated text \\
     $t_p^{d}$ & direct inverse prompt\\
     $t_p^{b}$ & block inverse prompt \\
     $t'$ & target text for inverse prompting perplexity computation \\
     \bottomrule
  \end{tabular}
\end{table}
Table ~\ref{tab:notations} displays notations used in this paper. We use $\mathcal{M}$ for generative language models and $\mathcal{D}$ for the probability distribution of tokens predicted by the model.   $\mathcal{D}$ is used to sample generations and compute perplexity scores. 
We use $t_p$ for prompt text and $t_g$ for generated text. $t_p^{d}$ refers to inverse prompt using direct generation models, while $t_p^{b}$ refers to inverse prompt using block generation models. $t'$ is the target for perplexity score computation. 
}
\subsection{Inverse Prompting}
Inverse prompting~\cite{zou2021controllable} is a controllable generation method that prompts pre-trained generative models under an inverse way using the inverse representation of natural language. The perplexity of the original prompt under the inverse form is computed and used as a scorer for beam search. The method greatly improves the generation quality of generative pre-trained models. 

The problem of text generation is modeled as generating text $t_g$ given the prompt text $t_p$, where both $t_p$ and $t_g$ are sequences of tokens.

A language model $\mathcal{M}$ takes prompt sequence $t_p$ and outputs a probability distribution of the next token $\mathcal{M}(t_p)=\mathcal{D}(tokens)$ over all available tokens.

For generation texts longer than a single token, the model generates in an auto-regressive way, sampling a token from $\mathcal{D}$ and appends the token after the prompt sequence $t_p$.

To improve consistency between prompt and generated text, inverse prompting aims to maximize conditional probability $p(t_p|t_{g})$, the probability to reconstruct the prompt given the generated context.
\beq{
\label{eqn:inv-1}
\max_{t_g}f(t_{g} | t_p)=\max_{t_g}\log p(t_p|t_g).
}
 $p(t_p|t_{g})$ cannot be directly achieved, inverse prompting estimates this by inverse transformation of natural language. 
\beq{
\label{eqn:inv-2}
\max_{t_g}f(t_{g} | t_p)=\max_{t_g}\log p(t'|t_p^{d}),
}
In equation ~\ref{eqn:inv-2}, the target text $t'$ and direct inverse prompt $t_p^{d}$ are inverse representations transformed from $t_p$ and $t_g$ using the inverse form of natural language.  
Some examples of inverse expression transformation are listed in Table ~\ref{tab:inv}. 

In summary, inverse prompting basically improves the quality of the generated text by offering a scorer that helps the model determine which generation is better. 

\subsection{Block Inverse Prompting}
\method can be viewed as a broader implementation of inverse prompting under block generative models($\mathcal{M}_b$), models that are able to generate intermediate text given previous and following text. 

Given prompt sequence $t_p$, block position $b$, already generated text $t_g$, the model outputs a probability distribution of the next token $\mathcal{M}_b(t_p,t_g,b)=\mathcal{D}(tokens)$ over all available tokens. Sampling from $\mathcal{D}$ and append the sampled token to $t_g$ we can generate intermediate text of any length in an auto-regressive way using block generative models. 

For block generative models, instead of relying on inverse transformation of natural language, $p(t_p|t_g)$ in equation ~\ref{eqn:inv-1} can be directly computed by simply mask $t_p$ and prompting the model with $t_g$. 
\begin{figure*}
    \includegraphics[width=\textwidth]{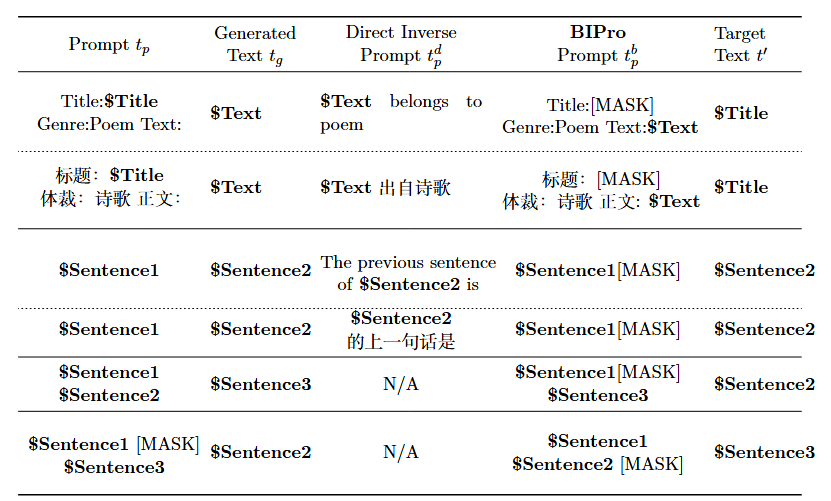}
    \caption{\label{tab:inv} Examples of formats used in direct inverse prompting and \method. \method directly masks the prompt and evaluate the perplexity under block generative models, skipping the inverse transformation process in direct inverse prompting.  }
\end{figure*}

Figure ~\ref{tab:inv} summarizes formats of prompts and targets for inverse prompting and \method used in poem generation. As can be seen, instead of using an inverse transformation in natural language, with the help of block generative models, \method is more direct. It avoids indistinct expressions of meanings in inverse transformation, and can handle conditions that are hard to construct natural inverse prompts, such as evaluating sentences in the middle of two sentences.   
\vspace{-0.1in}
\subsection{Constrained Generation}
\label{sec:constrain}
 In constrained generation, the generated text shall satisfy constraints. Constraints can be conspicuous. They may limit the vocabulary at some positions. Constraints can be inconspicuous. The usage of some words may temporarily satisfy the constraint while making it impossible for further text to lie within constraint. Handling with such constraints is more difficult than dealing with conspicuous constraints, as such dead ends are hard to detect in advance.
\SetKwInOut{Parameter}{Parameter}

\begin{figure}
\includegraphics[width=0.52\textwidth]{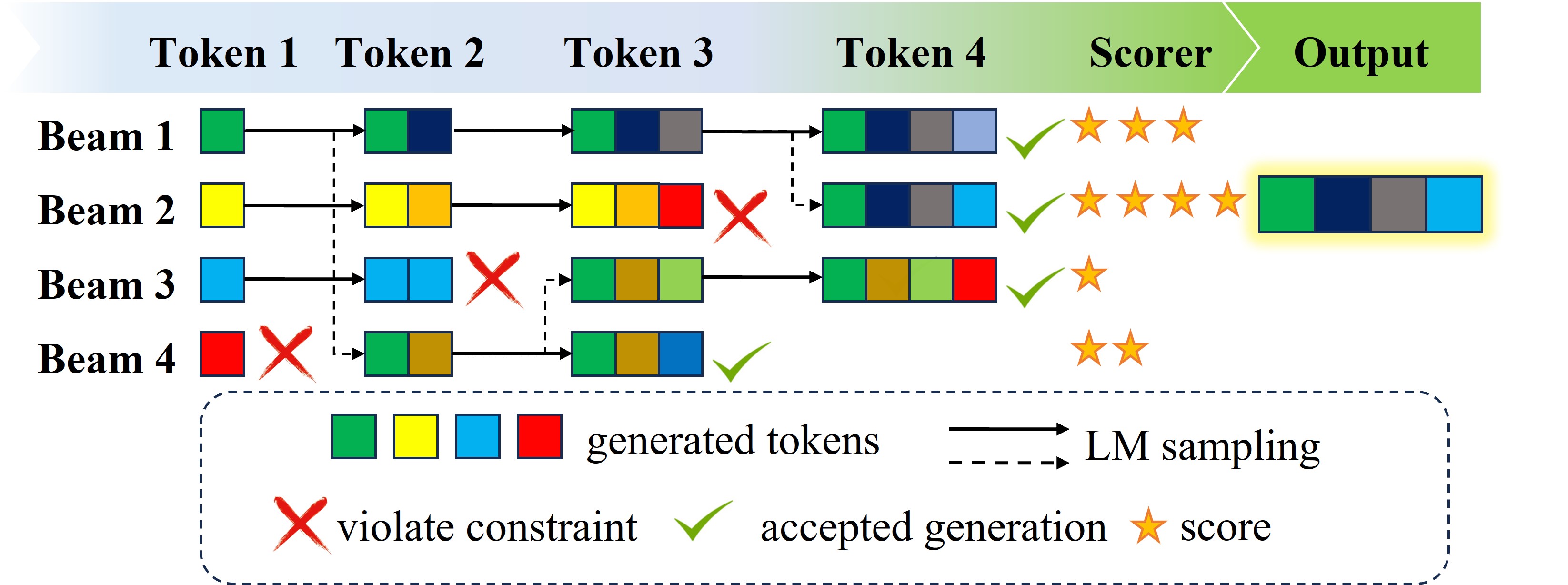}
\caption{\label{fig:constrainedgen} Beam-based constrained generation. Bad generation are replaced by good generations from other beams at each step. Finally, generations that satisfy constraints are scored and selected accordingly. }
\vspace{-0.12in}
\end{figure}
\begin{figure}
\includegraphics[width=0.52\textwidth]{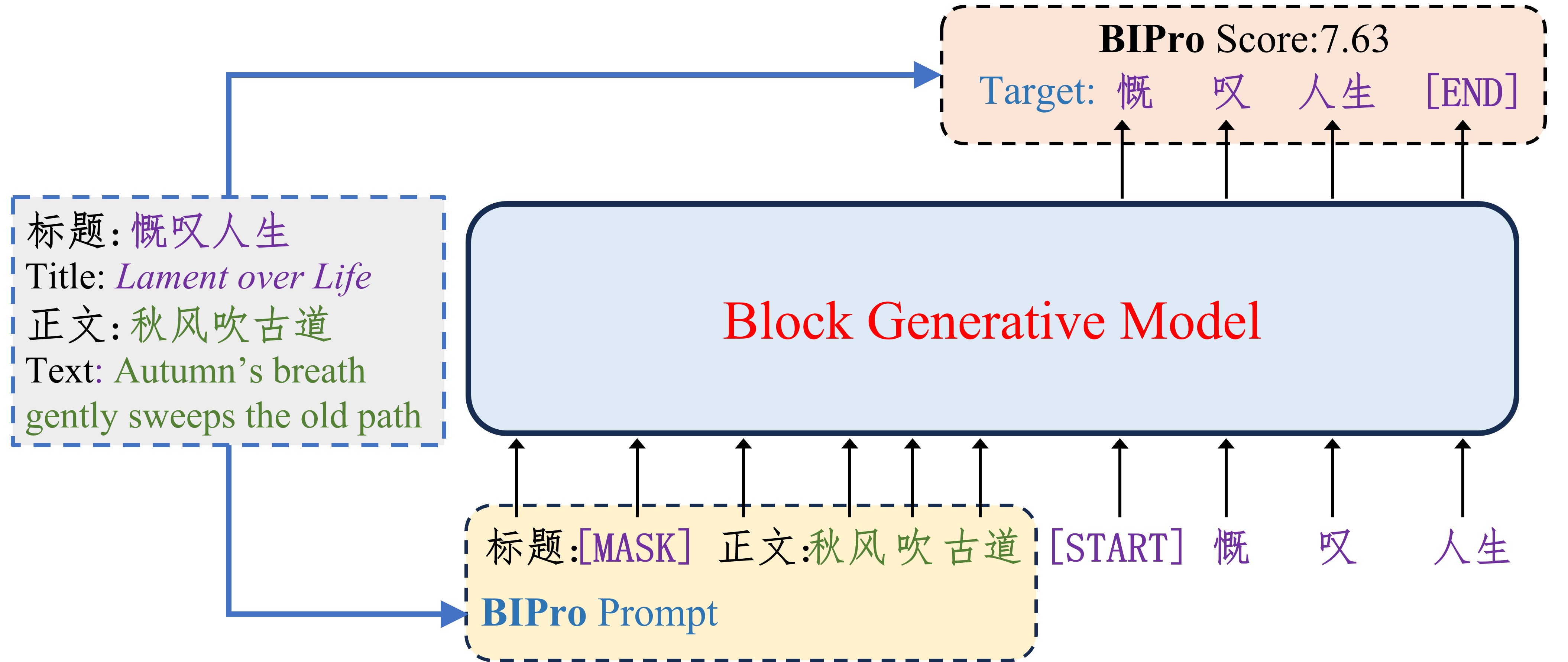}
\caption{\label{fig:scorer} \method \ scorer. The input is first transformed to \method prompt and target text, then \method prompt is fed into block generative model and the perplexity of the target text is used for scoring. }
\vspace{-0.12in}
\end{figure}
In \method, we use a search-and-evaluate strategy to generate poem sentences that satisfy the Pingshui constraint, illustrated in figure ~\ref{fig:constrainedgen}.  During generation, a number of beams is maintained. Beams violating constraints are replaced by secondary generations from other beams that still fit in constraints. By maintaining a population of generated texts, the strategy can overcome most of the dead ends of poem constraints. 

The generation process continues until all beams reaches an end. Eventually, all generations are evaluated by a scorer and the best beam is selected as the output generation. 

Figure ~\ref{fig:scorer} illustrates the \method scorer, the prompt and the generated text are transformed to \method prompt and target text according to Table ~\ref{tab:inv}. The \method prompt is fed into the block generative model, the perplexity of the target text is used as \method score. 
\subsection{\method Generation}

 Direct generative systems cannot revise texts. After subsequent texts are generated, they are unable to retrospectively alter the existing content. This rigidity contrasts with the human approach to text production, where rewriting, revising, and formatting are essential~\cite{seow2002writing}, particularly when crafting high-quality texts.
 Individuals often deliberate extensively, seeking for the optimal expression and making adjustments to the text they have already composed. Such flexibility is challenging for direct generative models, which generally lack the capability to modify earlier sections based on later ones.

Block generative models offer a solution by enabling the generation of intermediary text that considers both preceding and subsequent content, thus facilitating a writing process that more closely resembles human behavior.

\begin{algorithm}
\SetAlgoLined
\KwResult{Generated Poem $p$}
 \KwIn{Block generative model $\mathcal{M}_b$, input prompt $t_{p}$, constraint verifier $v$, \method scorer $s$}
 \Parameter{number of sentences $n$, maximal revise $m$}
 Initialize p=(),r=0\;
\For{$k \gets 1$ to $n$} {
     Generate sentence $p_k\leftarrow \mathcal{M}_b(t_p,p)$\;
     $p\leftarrow(p_1,...,p_k)$\;
     \If{k>1}{
      \tcc{\textbf{Revise}}
     $p_{k-1}^{'}\leftarrow\mathcal{M}_b(t_p,p/p_{k-1})$\;
     \If{$s(p/p_{k-1},p_{k-1}^{'})>s(p)$}
     {
     $p_{k-1}\leftarrow p_{k-1}^{'},p\leftarrow(p/p_{k-1},p_{k-1}^{'})$\;
     }
    }
 }
\While{($r<m$) and ($p$ changes in the last rewrite)}{
\tcc{\textbf{Rewrite}}
 \For{$k \gets 1$ to $n$} {
     $p_{k-1}^{'}\leftarrow\mathcal{M}_b(t_p,p/p_{k-1})$\;
     \If{$s(p/p_{k},p_{k}^{'})>s(p)$}
     {
     $p_{k}\leftarrow p_{k}^{'},p\leftarrow(p/p_k,p_k^{'})$\;
     }
    }
    $r\leftarrow r+1$\;
 }
 Output final poem $p$.
 \caption{\label{algo:bipro} \method Generation}
\end{algorithm}
 In this study, we introduce two methods, revise and rewrite. Revise refers to subtle and immediate modification of a sentence once the subsequent sentence has been produced.  
Rewrite refers to involves more substantial changes and is undertaken after the entire text has been generated. 
Figure~\ref{fig:poemgen} illustrates the \method constrained generation framework using an example of generating traditional-form Chinese poem under title ``\textit{Lament over Life}". Each sentence is generated using the constrained generation method described in the previous subsection. We revise each sentence immediately after its subsequent sentence is generated. We mask that sentence and prompt the model to generate a new one, then compare the new poem with the original one using \method scorer. We replace the original sentence with the new one when \method scorer gives it a better score. In the example of Figure ~\ref{fig:poemgen}, two sentences are revised during the initial generation.

Following the generation of a complete poem, we systematically rewrite each sentence by masking it and prompting the model with the remaining text, replacing them if the new generation is better in \method score. Such rewriting process can be cycled for multiple rounds until the model can no longer offer better expressions for any sentences of the poem, or the number of rounds achieves the set limit.
In the case of Figure ~\ref{fig:poemgen}, the poem is rewritten for 5 rounds.

The process of poem generation under \method framework is also described in Algorithm ~\ref{algo:bipro}.

\hide{
\subsection{Enhanced Attention}
Attention is the core part of Transformers~\cite{vaswani2017attention}, which can be described as mapping queries and keys of tokens to measure their importance and gathered their values together using a weighed sum according to the importance, as shown in equation ~\ref{eqn:attention}. 

\beq{
\label{eqn:attention}
Attention(Q,K,V)=softmax(\frac{QK^T}{\sqrt{d_k}})V
}

$Q,K,V$ are queries, keys and values on a set of series packed together into a matrix, and $d_k$ is the dimension of token representation. In transformers, $Q,K,V$ are computed using a linear transformation of toke representation. 

To enhance the relativeness of certain token pairs, for example, the prompt and the generated texts when prompting generation texts, or the newly-generated text and the targeted prompt when doing inverse prompting, a straightforward way is to manually adjust the weights of those token pairs. Equation ~\ref{eqn:attention_mani} presents the method. 

\beq{
\label{eqn:attention_mani}
Attention(Q,K,V)=softmax(\frac{QK^T}{\sqrt{d_k}}+C)V
}

$C$ is a manually defined relativeness adjustment matrix, $C_{ij}$ represents the scale of attention shift of the $i^{th}$ token to the $j^{th}$ token. $C$ is set initially to 0, when we want to emphasize some certain tokens, we set $C_{ij}=c,i\in t$, where $t$ is the set of the emphasized token, while $c$ is the scale of the enhancement.

}

\vspace{-0.07in}
\section{Experiments}\label{sec:exp}
\vspace{-0.07in}
Most of the leading language models are direct generative models. 
Block generative models are rare.  Currently the best open-source block generative model may be GLM-10B~\cite{du2022glm} and GLM-130B~\cite{zeng2022glm}. We use the open-sourced Chinese version of GLM-10B~\footnote{\url{https://github.com/THUDM/GLM}} as our base model. Detailed implementations are described in Appendix.

To evaluate the quality of poems generated by \method framework, we organize two human review challenges. Reviewers in these challenges are amateur poets associated with universities or local poetry clubs. They are experienced in crafting traditional-form Chinese poetry.
\vspace{-0.1in}
\subsection{Experiment Settings}
 In each challenge, a number of titles are given to different poem generation systems.
 The resulting poems, authored anonymously to ensure impartiality, are then presented to the reviewers. To facilitate fair comparison, poems sharing the same title are grouped and provided in random orders.

Reviewers shall rate the poems based on four aspects.
\vspace{-0.1in}
\begin{itemize}
\setlength{\itemsep}{-4pt}
\item \textbf{Format}, how well the poem fit into the constraints fluently and euphonically. 
\item \textbf{Informativeness}, amount of useful information contained in the poem.
\item \textbf{Relevance}, how well the poem suits the given title. 
\item \textbf{Aesthetic}, artistic conception of the poem. 
\end{itemize}
\vspace{-0.1in}
Reviewers are also required to rate an overall score , and how they think others may score for each poem.

\vpara{Open-domain Poem Generation} In this challenge, 42 titles suggested by different reviewers are gathered together and passed to 6 different poem generation systems: GPT-4~\cite{achiam2023gpt}, GLM-4, Baidu Poetry Helper, Yusheng~\cite{ma2023yu}, Shisanbai, and \method for poem generation. Reviewers shall review 6 different poems for each title. 

\vpara{Parallel Poem Generation} In this challenge, titles of 87 human-created traditional-form Chinese poems from \textit{Daily Poem} section of \textit{China Poetry} website~\footnote{\url{https://www.zgshige.com/}} are passed to 3 different poem generation systems, GPT-4, \method and GLM-10B direct generation. For each poem the systems shall generate poems with exactly the same format. Human poems are also included in the evaluation, resulting in a total of 4 poems for each title. 

Details of the implementations of \method and baseline models are included in Appendix.

All data used in the human review challenges do not contain personally identifying info or offensive content.
\hide{
\subsection{Baselines}
In this subsection we introduce the experimental settings for baselines. 

\vpara{Yusheng} Yusheng~\cite{ma2023yu} is the best published domain-specific traditional-form Chinese poem generation system  focus on traditional-form Chinese poem generation. It is a GPT-2 model trained on more than 1 million traditional-form Chinese poems, and is much better than previous systems like Jiuge~\cite{zhipeng2019jiuge} according to its evaluation. For each title, we randomly choose one of the four formats (``5-Jueju",``7-Jueju", ``5-Lvshi",``7-Lvshi") and input the title to the system on its public website.~\footnote{\url{https://yusheng.cis.um.edu.mo/}}

\vpara{Shisanbai} Shisanbai~\footnote{\url{https://www.aichpoem.net/\#/shisanbai/poem}} is a domain-specific traditional-form Chinese poem generation system. It is known for high-quality poem generation. The initial version of Shisanbai uses around 1 million traditional-form Chinese poems, and further updated details are unknown.
The free version of Shisanbai can only generation poems under a   limited range of titles.  We purchase VIP of it and generate poems using the VIP version. 

\vpara{GPT-4} 
OpenAI's GPT-4 has the ability to generate traditional-form Chinese poems. GPT-4-1106-Preview is better than previous versions in traditional-form Chinese poem generation. To increase the success rate and generation quality, we use few-shot prompt that includes 4 most famous traditional-form Chinese poems and ask GPT-4-1106-Preview to generate poems. Details of the few-shot prompt we use are in Appendix.  

\vpara{GLM-4}
GLM-4~\footnote{\url{https://chatglm.cn/}} is a large generative system published by Zhipu.AI. Its Chinese ability is close to GPT-4's and is also able to generate traditional-form Chinese poems given prompt. We use the same few-shot prompt as GPT-4 to guide it to generate poems that suits the format. 

\vpara{Baidu Poetry Helper}
Baidu Poetry Helper ~\footnote{\url{https://chat.baidu.com/bot?appId=80f66d44ac194b2684de766fd3d9b990&source=container}} is a helper released by Baidu. It is likely a fine-tuned version of ERNIE Bot on general poems. It is not specified to traditional-form Chinese poems so we use the few-shot prompt and additional instructions like GPT-4 to guide it to generate poems that suits the format. 

\vpara{Direct Generation}
We also compare with direct constrained generation of GLM-10B. In this case we still use the same prompt and beam-based constrained generation described in figure ~\ref{fig:constrainedgen}. However, \method scorer is not used and we simply select the output of the first beam. Revising and rewriting are also not used.  
}

\begin{table*}[htbp]
  \centering
  \vspace{-0.15in} 
  \begin{threeparttable}
  \begin{tabular}{cccccccc}
  \toprule
      \makecell[c]{Challenge} & \makecell[c]{Generation\\System} & \makecell[c]{Format\\(1-5)} & \makecell[c]{Infomativeness\\(1-5)} & \makecell[c]{Relevance \\(1-5)}  & \makecell[c]{Aesthetics \\(1-5)}& \makecell[c]{Overall\\(1-10)} & \makecell[c]{AR\\(1-10)} \\
    \midrule
  \multirow{6}*{\makecell[c]{Open-\\domain \\ Poem \\Generation}}&  Yusheng&3.43 &3.24&2.40&\textbf{3.08}&4.62&4.66\\
   & Shisanbai &\textbf{3.68}&3.34&2.94&3.01&5.13&5.16 \\
    & GPT$-$4 & 2.50 & 3.19 & 3.71 & 2.67 & 4.79 & 4.60\\
     &GLM$-$4 & 2.58 & 2.95 & 3.70 & 2.46 & 4.72 & 4.40\\
     &Baidu & 2.66 & 3.17 & \textbf{3.73} & 2.51 & 4.76 & 4.70 \\
      &\method  &3.26& \textbf{3.42} & 3.30 & 2.93 & \textbf{5.27} & \textbf{5.22} \\
  \midrule
  \multirow{4}*{\makecell[c]{Parallel \\ Poem \\Generation}} & Direct &2.90&2.99&2.70&2.63&4.65 & 4.37 \\
  &  GPT$-$4 & 2.64&3.00&\textbf{3.59}&2.56&4.98 & 4.86\\
   &   \method&\textbf{3.26}&\textbf{3.33}&3.34&\textbf{2.92}&\textbf{5.54} & \textbf{5.43} \\
  & \textit{Daily Poem} &3.55&3.59&3.84&3.30&6.37 & 6.42\\
  \bottomrule
  \end{tabular}
  \end{threeparttable}
    \caption{\label{tab:exp1}Experimental results of poem generation challenges. Best scores of AI generation systems are bolded.}
  \vspace{-0.1in}
\end{table*}
\vspace{-0.1in}
\subsection{Experimental Results}
Table ~\ref{tab:exp1} displays the results of the two human review challenges. We provide averaged detailed scores and overall scores. Recognizing that reviewers vary in their evaluation criteria, we also collect their predictions for scores others may rate to each poem. Utilizing this data, we apply \textit{Answer Ranking} (AR) ~\cite{kong2022eliciting} and calculate an AR score for each poem. We also provide the averaged AR score for each method.

In open-domain poem generation challenge, 
due to the lack of domain-specific training for base model, 
\method is not as good in satisfying format constraints or generating aesthetic poems as domain-specific poem generation systems Yusheng and Shisanbai. 
Using a less powerful base model of GLM-10B, the generated poems from \method are also not as related to the given title as those generated by leading second-generation generative systems GPT-4 or GLM-4. 
However, the \method framework excels at guiding GLM-10B-Chinese in producing poems that balance between format, aesthetics, informativeness, and title relevance. The poems from \method received the highest overall average score(5.27) and AR score(5.22) from reviewers, outperforming all other methods on the open-domain titles the reviewers proposed.  

 In parallel generation challenge, the overall average of direct generation using GLM-10B stands at 4.65,  with the AR score lagging even further behind at 4.37.
  In comparison, poems crafted by GPT-4 outperform those from GLM-10B with a higher average overall score of 4.98 and AR score of 4.86. It is important to note that this comparison might understate the true disparity in quality, as the GLM-10B output may include instances replicated from well-known poems. An example is offered in Appendix.  
The \method framework markedly enhances the caliber of poems generated using the base model of GLM-10B-Chinese, elevating the average overall/AR score to 5.54/5.43.

In every measure, human poems from \textit{Daily Poem} maintain their supremacy, boasting an average overall score of 6.37 and AR score of 6.42. Although \method represents the present state-of-the-art in automated traditional-form Chinese poem generation, using GLM-10B as the underlying model, its output has yet to match the nuanced artistry of human poetry.
\vspace{-0.1in}
\subsection{Case Study}
\label{sec:case}
\begin{figure}[h]
\vspace{-0.1in}
    \includegraphics[width=0.48\textwidth]{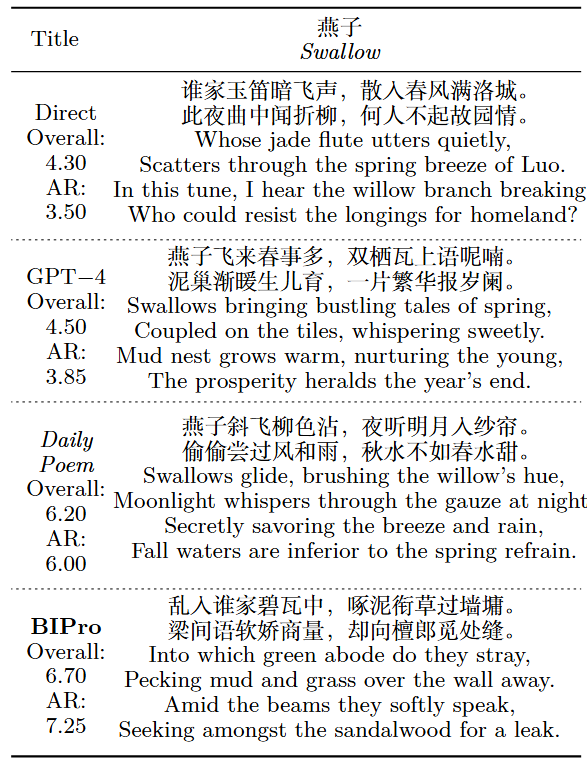}
    \vspace{-0.1in}
    \caption{\label{tab:tcp-swallow}A representative case in parallel poem generation challenge. }
    \vspace{-0.15in}
\end{figure}

Figure ~\ref{tab:tcp-swallow} shows an instance in which poem generation systems are tasked with generation a ``7-Jueju" poem under the title \textit{Swallow}, paralleling an existing human poem on ~\textit{Daily Poem}. 

Direct generation copies a famous ancient poem with another title. Such an approach yields verses that are somewhat relevant and get modest scores, but it lacks originality, suggesting the generation ability of GLM-10B without \method is weak.

GPT-4 outperforms this approach by producing fresh and title-appropriate content. However, its creation tends to be lack in aesthetics and has critical error that wrongly refers spring to the end of the year. It only gets slightly higher scores.

In contrast, the human poem from \textit{Daily Poem} is novel and exquisite. It presents a good mixture of the title concept and nature concepts like night, moon, breeze, rain, autumn and spring. This poem gets 6.20 average score and 6.00 AR score, which is the typical level of short-listed human poems in \textit{Daily Poem}. 

The generation from \method is even better. Through rounds of revising and rewriting, instead of directly using the title concept, the poem uses descriptions of behaviors to imply the existence of a swallow. It also expresses some deeper subtle human-like feelings. Reviewers rate this generation with 6.70 average score and 7.25 AR score, better than its human counterpart. 

The huge leap from direct generation to \method framework using the same base model GLM-10B shows the immense power of \method framework on elevating qualities of generated texts on constrained generation tasks. More samples are listed in Appendix. 
\vspace{-0.07in}

\section{Conclusion} \label{sec:conclusion}
\vspace{-0.07in}

In this paper, we introduce \method framework, an innovative constrained generation framework that leverages the capabilities of block generative models by iteratively revise and rewrite the generated texts using the model itself. The \method framework empowers these models to produce significantly improved texts within predefined constraints. 

Through human review, we have evidenced that \method enables a relatively modest block generative model, GLM-10B, to outperform stronger direct generative models as well as best domain-specific systems in the formidable arena of open-domain traditional-form Chinese poem generation, accomplished with zero-shot prompts and no additional domain-specific training. 


\clearpage
\newpage
\section*{Limitations}
\subsection*{Computational Complexity}
As shown in algorithm ~\ref{algo:bipro}, \method framework involves selection, scoring and iterative process during generation, which causes extra computational resources. To be precise, if generating a sentence uses $s$ tokens, the beam size in constrained generation is $k$, the target text has a length of $t$, then \method spends $k(t+s)$ tokens to generate a single sentence, $2nk(t+s)$ tokens to generate a full poem of $n$ sentences including revise, and $nk(m+2)(t+s)$ tokens for the full revise-rewrite process.

This computational cost is $O(mk)$ times more than direct generation. If we take a usual parameter set of $n=8, s=7, m=10, t=5, k=6$, then generating a poem spends around 7000 tokens, which is far more than 50 tokens for direct generation. The computational complexity may limit its usage at very large scales.

In real-world experiments, all poems can be generated within 1 minute time using a single A100 GPU.  Commercially, if we use GPT-4o mini($\$0.3/1m$ token)~\footnote{\url{https://openai.com/api/pricing/}} as reference for potential upgrades to larger models, the cost will be around $\$0.002$ per poem. 

As token costs for pre-trained language models decreasing rapidly due to technological development, controllable generation methods like this work that consumes more token for better generations may become increasingly useful.

\subsection*{Lack of Automated Benchmarks}
\method aims at improving generation quality at constrained generation. Evaluating qualities of texts under specific constrained generation tasks is hard, especially under artistic settings like poetry. It seems impossible to create automated benchmarks for poem quality evaluation and we have to rely on human reviewers. 

We choose the task of traditional-form Chinese poem generation mainly because of its well-known and relatively easy access to sufficient reviewers. For other potential constrained generation tasks, we believe \method can still work given proper constraint requirements but the improvement may be hard to review.  

We acknowledge that human reviewers may not be perfect, and adopt various further measures to increase the accuracy of our experiments. See Appendix for details. 

\subsection*{Potential Abuse}
 \method can largely improve quality of texts under constrained generation situations. When applied on potential better block generative models in the future, it may surpass the best humans on poem generation or other constrained generation tasks in double-blind human evaluations.
 
Being a potentially public available high quality and low cost constrained generation method, there may be abuse. This method may help improve qualities of negative content when constraints are set in a negative manner.
For example, creating various cute slogans to promote bad things, using poetry for sarcasm, or other abusive situations. 

Improving qualities for AI generations is always a double-edged sword. Powerful generative AIs can make the world better if used positively. They may worsen the world if used negatively. To maximize benefits and minimize side effects, appropriate supervision may be needed.

\newpage
\bibliography{reference.bib}

\clearpage
\newpage
\appendix
\section{Appendix}
\subsection{Implementation Details}
\subsubsection{Base Model}
We use \textit{SwissArmyTransformer}~\footnote{\url{https://github.com/THUDM/SwissArmyTransformer}} implementation of the open-sourced Chinese version of GLM-10B~\footnote{\url{https://github.com/THUDM/GLM}} as our base model. It is a block generative model with 9.87 billion parameters trained on general modern Chinese text dataset Wudaocorpora~\cite{yuan2021wudaocorpora}. We do not apply any additional fine-tuning so its knowledge on traditional-form Chinese poem is limited to its pre-train dataset of Wudaocorpora. 

GLM-10B isn't good at direct generation. Most benchmarks don't include it for its poor performance. The only benchmark including GLM-10B is the common knowledge benchmark M3KE~\cite{liu2023m3ke}. The benchmark evaluates the open-domain Chinese common knowledge for models by asking them common knowledge questions of all fields.  
GLM-10B gets $19.7\%$ overall accuracy, much worse than ChatGLM-6B's $23.6\%$ or GPT-4's $63.8\%$, detailed performance on M3KE benchmark is displayed in table ~\ref{tab:m3ke}.
\begin{table*}
   \centering
   \small
   \begin{tabular}{cccccc}
    \toprule
      Domain & GLM-10B & ChatGLM-6B &  GPT-4 \\
      \midrule
      Arts\&Humanities & 0.180 & 0.246 &0.588 \\
      Social Sciences & 0.229 & 0.267 & 0.676 \\
      Nature Sciences & 0.219 & 0.168 & 0.623 \\
      Others & 0.150 & 0.263 & 0.665 \\
      Average & 0.197 & 0.236 & 0.638 \\
     \bottomrule
  \end{tabular}
   \caption{\label{tab:m3ke} Performance(Accuracy of multiple choice) of different models on Chinese common knowledge benchmark M3KE~\cite{liu2023m3ke}.}
\end{table*}

\subsubsection{Generation with Constraint}
We use the beam-based constrained generation strategy described in section ~\ref{sec:constrain} to generate multiple candidates for each sentence, and use the perplexity of the target given \method prompt as the scorer.

In our application, we use the weighted sum of \method score for the title and \method score for another poem sentence of the generated sentence as scorer, and use a beam size of 6. We use \method score for the previous sentence during the generation phase, and \method score for the following sentence in revising. The ``match" sentence, which is the next sentence of odd sentences and previous sentence for even sentences is used for \method score in rewriting. We set the maximal round of rewriting to 20. In practice, poems are usually generated within 1 minute time using a single A100 GPU.

 We use a Pingshui format verifier to ensure generations on all beams follow the Pingshui constraint. 

In our experiments, we use zero-shot prompts that only tells the model to produce a poem without offering any examples.

\subsubsection{Compared Methods}
In this section we discuss the implementation details of compared methods. 

\vpara{Yusheng} Yusheng~\cite{ma2023yu} is the best published domain-specific traditional-form Chinese poem generation system  focus on traditional-form Chinese poem generation. It is a GPT-2 model trained on more than 1 million traditional-form Chinese poems, and is much better than previous systems like Jiuge~\cite{zhipeng2019jiuge} according to its evaluation. For each title, we randomly choose one of the four formats (``5-Jueju",``7-Jueju", ``5-Lvshi",``7-Lvshi") and input the title to the system on its public website.~\footnote{\url{https://yusheng.cis.um.edu.mo/}} We don't use Jiuge as Yusheng is a higher-level substitute. 

\vpara{Shisanbai} Shisanbai~\footnote{\url{https://www.aichpoem.net/\#/shisanbai/poem}} is a domain-specific traditional-form Chinese poem generation system. The initial version of Shisanbai uses around 1 million traditional-form Chinese poems, details for the current version is unknown. 
The free version of Shisanbai can only generation poems under a limited range of titles.  We purchase VIP of it and generate poems using the VIP version. 

\begin{figure*}[htbp]
\includegraphics[width=\textwidth]{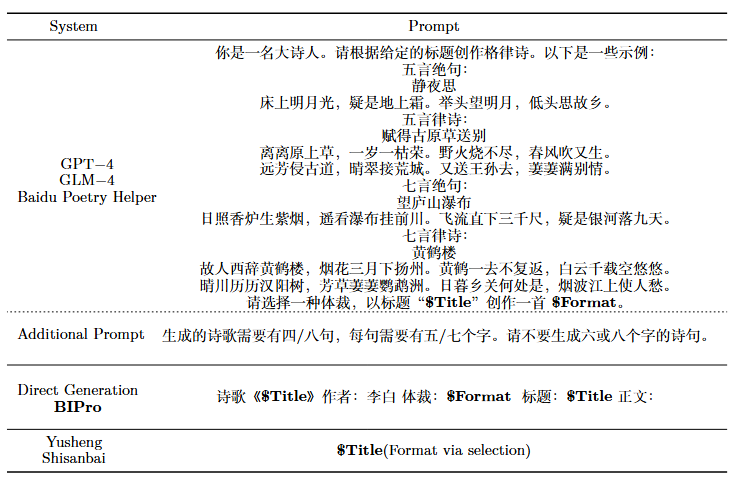}
  \caption{ \label{tab:tcp-prompt}Prompts used for different generation systems. Format requirement and title of the poem are denoted as \textbf{\$Format} and \textbf{\$Title}.}
\end{figure*}

\vpara{GPT-4} 
OpenAI's GPT-4 has the ability to generate traditional-form Chinese poems. GPT-4-1106-Preview is better than previous versions in traditional-form Chinese poem generation. To increase the success rate and generation quality, we use few-shot prompt for GPT-4-1106-Preview to generate poems. Sometimes the generation still does not fit the format requirement, on that case we add additional format guide prompts until it generates a poem that fit in the format. Table ~\ref{tab:tcp-prompt} displays the used prompt and additional prompt.  

\vpara{GLM-4}
GLM-4 is a large generative system published by Zhipu.AI. Its Chinese ability is close to GPT-4's and is also able to generate traditional-form Chinese poems given prompt. We use the same few-shot prompt/additional prompt as GPT-4 to prompt it. 

\vpara{Baidu Poetry Helper}
Baidu Poetry Helper~\footnote{\url{https://chat.baidu.com/bot?appId=80f66d44ac194b2684de766fd3d9b990&source=container}} is a helper released by Baidu. It is likely a fine-tuned version of ERNIE Bot on general Chinese poems. It is not specified to traditional-form Chinese poems so we use the few-shot prompt/additional prompt as GPT-4 to prompt it. 

\vpara{Direct Generation}
In parallel generation challenge, we also include direct generation from GLM-10B model as baseline to better display how much \method contributes. It is unlikely for GLM-10B model to generate poems directly under any prompt as it does not include poems in its training data. Hence the PingShui verifier and beam-based generation method in figure ~\ref{fig:constrainedgen} is used. But the \method scorer is not applied and the generation of the first beam that satisfy the PingShui constraint is selected. The generations do not experience revise or rewrites. 

\vpara{Daily Poem}
Being a well-known constrained writing type with rich culture, there exist quite a few populated traditional-form Chinese poem communities. \textit{China Poetry}~\footnote{\url{https://www.zgshige.com}} is one of them. Poets register and submit their poems on the website. Poems on the website may be modern form or traditional form. The website runs a \textit{Daily Poem}~~\footnote{\url{https://www.zgshige.com/mrhs/}} section that select best recent poems submitted by poets. Poets may be rewarded a small amount of money if their poems are selected. 
We focus on the traditional-form poems in the \textit{Daily Poem} section. We collect all of the traditional-form poems from May 2023 to December 2023 in the section for parallel poem generation challenge. 
Figure ~\ref{tab:tcp-prompt} lists the prompts used for different generation systems. We use few-shot prompt for direct generative systems like GPT-4, GLM-4 and Baidu Poetry Helper. The few-shot prompt offers those systems information on formats of traditional-form Chinese poems so that they're more likely to generate well-formatted poems. 

There are four formats of traditional-form Chinese poems, ``5-Jueju",``5-Lvshi",``7-Jueju",``7-Lvshi", in open-domain poem generation challenge we don't limit the format and only inputs ``Poem" as format. In parallel poem generation challenge we also limits the format to be concurrent to the human poem. 

Sometimes these systems don't generate poems that fit into those four types, on that case we point out the error and try to guide it to generate a poem fit in the format by using additional prompt. We use the first generation result that fits in the format as the candidate.

Domain specific systems like Yusheng and Shisanbai have built-in format regulators, so we simply choose the format and input the title.
 In direct generation experiment, we uses the beam-based constrained generation method to generate each sentence to ensure the result fits the constraint, and simply select the output of the first beam instead of using \method scorer to select a best beam.
\begin{table*}[ht]
    \centering
    \vspace{-0.1in}
    \begin{tabular}{ccccc}
    \toprule
        Challenge & Titles & \makecell[c]{Compared \\Methods} & Reviewers & \makecell[c]{Scores\\ Collected}\\
         \midrule
         \makecell[c]{Open-domain \\poem generation} & 42 &6 & 10   & 2,520 \\
         \cdashline{1-5}
         \makecell[c]{Parallel \\poem generation} & 87 &4 & 10 & 3,480\\
    \bottomrule
    \end{tabular}
    \caption{ \label{tab:humaneval}Statistics for two human evaluation experiments.  }
    \vspace{-0.1in}
\end{table*}
Table ~\ref{tab:humaneval} displays the statistics for the two human evaluation challenges.

\subsection{More Samples of Generations}
\begin{figure*}[h]
    \includegraphics[width=\textwidth]{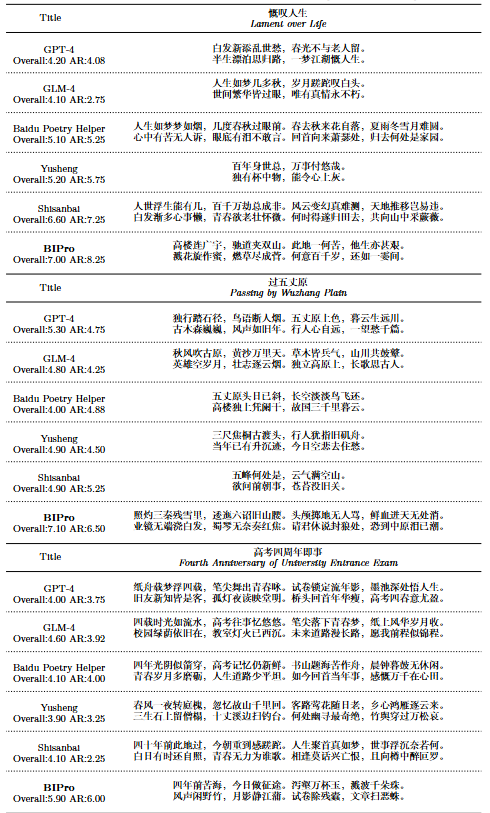}
    \caption{\label{tab:tcp-life}Representative poems in open-domain poem generation challenge.  }
\end{figure*}

\begin{figure*}[h]
    \includegraphics[width=\textwidth]{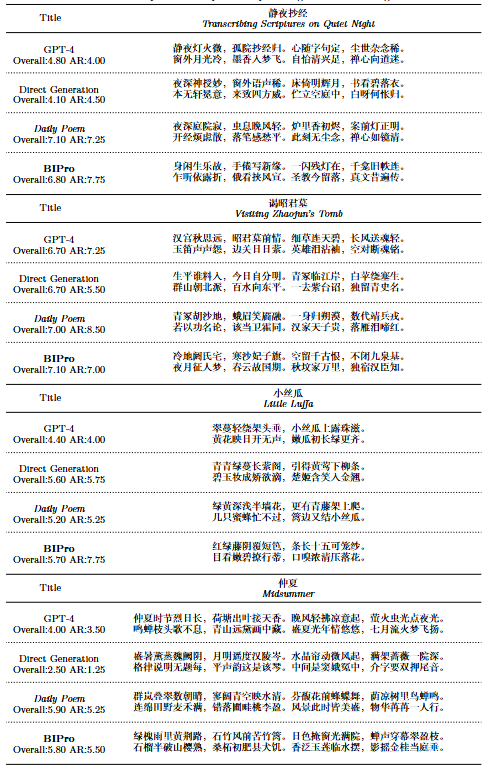}
    \caption{\label{tab:tcp-para}Representative poems in parallel generation challenge. }
\end{figure*}

In section ~\ref{sec:case} we display an representative case to show the generation of different systems. Figure ~\ref{tab:tcp-life} and Figure ~\ref{tab:tcp-para} offer more examples. The overall scores and Answer Ranking scores of those examples are also attached. 
Representative poems in parallel generation challenge.

\subsection{Human Evaluation Details}
We recruit poets from university or local poetry groups to evaluate the poems. All reviewers are amateur poets that have composed some traditional-form Chinese poems. A total of 12 poets are included in the review, 2 of them only participated in the evaluation of open-domain poem generation challenge, 2 only participated in the evaluation of parallel poem generation challenge. The remaining 8 reviews both challenges.  

The human evaluation is conducted on an online platform. For open-domain traditional-form Chinese poem generation challenge, the whole task of evaluating  prompts is divided into 6 sub-tasks, and in each sub-task, the evaluator is required to score 6 poems composed by different AI systems for 7 titles, like an online questionnaire.

For parallel poem generation challenge, the whole task of evaluating  prompts is divided into 8 sub-tasks, and in each sub-task, the evaluator is required to score 4 poems composed by 3 different AI systems and human for 11 titles, resulting in 88 titles. However, one of the title is controversial for scoring so we removed it and take evaluation results of the remaining 87 groups for analysis. 

The evaluation does not necessarily need to be finished at once. People can login and logout, change their answers for already completed problems, or continue evaluation from their current points freely. They only need to ensure that all evaluation questions have been answered. 

Each reviewer is paid 200 RMB for identical-title poem generation challenge and 150 RMB for open-domain traditional-form Chinese poem generation challenge. The top 3 reviewers in estimating scoring of others get a bonus of 200,100 and 50 RMB in both tasks.

Statistics of human evaluation experiments are presented in Table ~\ref{tab:humaneval}.

We also provide 
variance of scores of the two challenges in Table ~\ref{tab:var}.

\begin{table*}
  \centering
  \vspace{-0.1in}
  \caption{\label{tab:var}Average and variance of scores by reviewer in poem generation challenges. } 
  \begin{threeparttable}
  \begin{tabular}{ccccccc}
  \toprule
      \makecell[c]{Challenge} & \makecell[c]{Generation\\System} & \makecell[c]{Format\\(1-5)} & \makecell[c]{Infomativeness\\(1-5)} & \makecell[c]{Relevance \\(1-5)}  & \makecell[c]{Aesthetics \\(1-5)}& \makecell[c]{Overall\\(1-10)}  \\
    \midrule
  \multirow{6}*{\makecell[c]{Open-\\domain \\ Poem \\Generation}}&  Yusheng&3.43 $\pm$0.64 &3.24$\pm$0.53 &2.40$\pm$0.49&3.08$\pm$0.62&4.62$\pm$1.24\\
   & Shisanbai &3.68$\pm$0.77&3.34$\pm$0.49&2.94$\pm$0.59&3.01$\pm$0.45&5.13$\pm$1.18 \\
    & GPT$-$4 & 2.50$\pm$0.55 & 3.19$\pm$0.57 & 3.71$\pm$0.64 & 2.67$\pm$0.43 & 4.79$\pm$0.97 \\
     &GLM$-$4 & 2.58$\pm$0.54 & 2.95$\pm$0.52 & 3.70$\pm$0.66 & 2.46$\pm$0.46 & 4.72$\pm$0.88 \\
     &Baidu & 2.66$\pm$0.50 & 3.17$\pm$0.55 & 3.73$\pm$0.67 & 2.51$\pm$0.37 & 4.76$\pm$0.80 \\
      &\method  &3.26$\pm$0.73& 3.42$\pm$0.59 & 3.30$\pm$0.59 & 2.93$\pm$0.56 & 5.27$\pm$1.07  \\
  \midrule
  \multirow{4}*{\makecell[c]{Parallel \\ Poem \\Generation}} & Direct &2.90$\pm$0.49&2.99$\pm$0.46&2.70$\pm$0.58&2.63$\pm$0.45&4.65$\pm$1.17  \\
  &  GPT$-$4 & 2.64$\pm$0.51&3.00$\pm$0.42 &3.59$\pm$0.51&2.56$\pm$0.46&4.98$\pm$0.86 \\
&\method&3.26$\pm$0.50&3.33$\pm$0.47&3.34$\pm$0.51&2.92$\pm$0.37&5.54$\pm$0.88  \\
  & \textit{Daily Poem} &3.55$\pm$0.45&3.59$\pm$0.37&3.84$\pm$0.41&3.30$\pm$0.31&6.37$\pm$0.66\\
  \bottomrule
  \end{tabular}
  \end{threeparttable}
  \vspace{-0.12in}
\end{table*}

\subsection{Answer Ranking(AR) Score}
Although all human reviewers are proficient poets, they may have different understandings and evaluate each poem differently. Human-written or AI generated traditional-form Chinese poems usually have subtle metaphors that is uneasy to fully understand. On that case simply averaging scores from each reviewers may not reflect the actual level.  

To tackle this problem, ~\cite{kong2022eliciting} proposes the \textit{Answer Ranking} method. The core idea of this method is that reviewers may have different levels of thinking, and the levels can be extracted by asking each reviewer to predict how other reviewers' choice.
\begin{equation}
\label{eqn:AR}
\pi^*=\arg\max_{\pi}\sum_{\pi_i\leq\pi_j}M_{i,j}^2
\end{equation}
Suppose the reviewers' own choice and predicted choice forms a matrix $M$, $M_{i,j}$ is the number of people who choose $i$ and predicts $j$, then 
the best answer ranking is computed by equation ~\ref{eqn:AR}. $\pi$ refers to a ranking, and our goal is to maximize the L2-norm of the people who chooses a higher ranking answer than his predicted other people's choice. 

In our experiments, as scores are continuous, we only allow two types of rankings, $\pi=(n,n+1,n-1,n+2,n-2,...)$ which corresponds to score $n+0.25$, and $\pi=(n,n-1,n+1,n-2,n+2,...)$ which corresponds to score $n+0.75$. We use the score corresponding to the best $\pi^*$ as the AR score of a poem. If there exists multiple best rankings, we average the scores correspond to those rankings. So the actual range for AR score is 1.25 to 9.75. 

\subsection{Pingshui Constraint}
Pingshui is the widely-accepted constraint system for traditional-form Chinese poems. Under Pingshui, each  Chinese characters belong to one of the two major categorizes, ``Ping" and ``Ze" based on their ancient pronunciation. Characters further divided to 106 sub-categorizes for rhyming propose.

The fundamental constraints of traditional-form Chinese poems are listed as following:
\begin{itemize}
\item[1] A poem must contain 4 or 8 sentences. Poems with 4 sentences are called ``Jueju" while poems with 8 sentences are called ``Lvshi".
\item[2] Each sentence must have 5 or 7 characters, the number of characters in different sentences of the same poem must be the same.
\item[3] Odd sentences shall end with a ``Ze'' character, even sentences shall end with a  ``Ping'' character, except for the first sentence, which can end freely.  
\item[4] The last characters of the even sentences shall rhyme. They shall be different characters and they shall all belong to the same one of the 106 sub-categorizes of Pingshui. If the first sentence ends up with ``Ping", it shall also satisfy this constraint. 
\item[5] For each sentence, the 2nd character shall belong to a different major category than the 4th character. If the sentence has a length of 7, then the 6th character shall belong to the same major category of the 2nd character.
\item[6] The 2nd characters of even sentences shall belong to a different major category than the 2nd character of its previous sentence. The 2nd characters of odd sentences shall belong to the same major category than the 2nd character of its previous sentence, except for the first sentence.
\item[7] The last 3 characters of any sentence shall not belong to the same major category of either ``Ping" or ``Ze".
\item[8] If a sentence ends up with ``Ping", then other ``Ping" characters in this sentence shall not have both their previous and next characters being ``Ze".
\end{itemize}

In this paper, the format verifier verifies each sentence based on these 8 constraints. In PingShui, some characters may have multiple pronunciations under different context, in such cases, the verifier will pass if any one of the pronunciations of these characters meet the constraint. 

For GPT-4, GLM-4 and Baidu Poetry Helper, it is very hard for their generations to meet all of the 8 constraints. We only check if the generation meets the first 2 constraints and repeatedly prompting them until they produce a poem that satisfies the two constraints. 

\subsection{Reproducability}
The code and data of this paper are open-sourced on github~\footnote{\url{https://github.com/xz-keg/BiPro}}. The repo includes the poem generation code, the review data of two experiments and the result analyze code.   

This work is also online on ChatGLM website as a tool Shiyun Zhineng.~\footnote{\url{https://chatglm.cn/main/gdetail/672c837c8ba8cf3453de646c?lang=zh}}

\end{document}